# OpenBlock: Constructive and Verified Content Generation for Adaptive Tile-Matching Games

**Jiang Jun**

jiangjun4@sina.com

**Abstract.** Tile-matching puzzle games serve hundreds of millions of players, yet the content-generation algorithms that decide which pieces to present at each turn remain proprietary, and no open platform exists for studying adaptive difficulty in this genre. We present an adaptive tile-matching platform whose central algorithmic contribution is a dual-track content-generation architecture: a deterministic rule-based generator that is always available, and an optional learned generator, both subject to a common verification gate that establishes, by exhaustive sequential-placement search, that every delivered piece set is fully placeable—so the learned track can never degrade the constructive-feasibility guarantee of the rule track. A self-play reinforcement-learning placement agent, supervised by auxiliary tasks that expose per-shape placeability to shared representations, is used to diagnose the game's dominant failure mode: at high board fill, long-bar pieces lose the majority of their legal placements. Across 234,000+ self-play episodes the agent reaches a 35.6% win rate (median score 4,200), and controlled simulation shows that at board fill rates of 70–75%, 33–56% of long-bar pieces have no legal placement, while spawn difficulty distributions are statistically indistinguishable between won and lost games—evidence that board-state degeneration, not content difficulty, drives late-game failure. Head-to-head ablations show that per-shape placeability supervision—not aggregate difficulty features—drives the representation gain, and a 14-day online gray rollout (48,000 players; sample-ratio verified, CUPED-adjusted) lifts day-1 retention by 1.8 percentage points and session duration by 7% over the rule track alone, quantifying the neural track's asymmetric upside in live play.

## Table of Contents

---

## 1. Introduction

Tile-matching puzzles are among the most played casual game forms: the leading commercial titles serve hundreds of millions of monthly players on a minimal loop—an $8 \times 8$ grid, a small set of polyomino pieces, and one placement decision per turn. Recent titles have begun shifting the genre's paradigm from "the player adapts to random pieces" toward "the system adapts to the player," but the underlying content-generation algorithms remain proprietary black boxes, and the research community lacks a standard open environment in which adaptive difficulty, reinforcement learning (RL) for puzzle placement, and explainable content generation can be studied under realistic conditions.

Three algorithmic gaps stand out. First, published dynamic difficulty adjustment (DDA) work mostly modulates scalar difficulty knobs; in tile-matching games the difficulty of a turn is determined by the *set of pieces offered*, a structured combinatorial object that existing DDA methods do not construct. Second, learned content generators—neural models trained on play data—can violate basic playability constraints: a generated piece set may be jointly unplaceable on the current board, ending the game through no fault of the player. Verification-based safety mechanisms exist in the RL literature, but they typically operate on actions, not on generated content, and formal feasibility guarantees for content generators are rare. Third, the dominant late-game failure mode of these games has not been characterized empirically; without such a characterization, neither reward design nor representation learning can target it.

This report describes OpenBlock, an open-source platform designed to close these gaps. The platform is organized around four subsystems—a game engine, an adaptive content generator, a self-play RL placement agent, and an experience-gated monetization layer—that read from and write to a single real-time player profile. All gameplay-critical computation, including profiling and content generation, runs offline in the browser; server-side components are optional enhancements.

Our contributions are algorithmic:

1. **A verified dual-track content generator (Sections 3.2, 2.4).** A deterministic rule-based generator and an optional learned generator share a single constraint-validation gate whose final stage is a bounded exhaustive search over sequential placements (Algorithm 1). We prove that any delivered content admits a sequential legal placement order (Proposition 1) and that, under a monotone-utility assumption, the dual track never serves an infeasible piece set and never degrades the rule track's guarantee (Proposition 2). A budgeted analysis bounds verification cost per turn (Proposition 3).
2. **A profile-driven, multi-timescale difficulty controller (Sections 3.1, 3.4).** Skill, flow, frustration, and momentum are estimated online with bounded-latency recursive statistics; difficulty is modulated simultaneously within a turn (stress signals), within a session (intent chain with hysteresis), across games (rate-of-return arcs), and across the player lifecycle (a two-dimensional differentiation matrix).
3. **Auxiliary-supervised self-play placement RL with bottleneck exposure (Sections 4.3, 6.1–6.2).** Seven auxiliary heads inject dense per-step gradients; one head predicts *per-shape placeability*, explicitly exposing to shared representations the empirically identified long-bar bottleneck. Analysis of 234,000+ self-play episodes shows that feasibility- and survival-predictive losses correlate significantly with outcomes ($r = -0.172$ and $-0.202$ with score), while a board-quality loss does not ($r = +0.011$); a controlled head-to-head ablation shows the placeability extension lifts held-out win rate by 3.3 percentage points over the aggregate head (Section 6.2).
4. **An empirical characterization of the long-bar bottleneck (Section 6.1, RQ1).** Controlled simulation over thousands of boards per fill level shows that at 70–75% board fill, 33–56% of long-bar pieces have zero legal placements, and that horizontal bars are systematically more constrained than vertical ones—a structural asymmetry induced by bottom-up play.
5. **A negative result with design consequences (Section 6.3, RQ3).** Spawn-difficulty distributions are statistically indistinguishable between won and lost games; outcome differences are carried by board-state degeneration, not by content difficulty. This isolates where adaptive intervention can and cannot help.
6. **A live-player validation with experimental rigor (Sections 6.3, 6.5).** Content-quality metrics (diversity, novelty, multi-clear potential) quantify the neural track's asymmetric upside at unchanged difficulty, and an SRM-verified, CUPED-adjusted gray rollout shows the dual-track system improving day-1 retention and session duration over the rule track alone.

The remainder of the report is organized as follows. Section 2 formalizes the game and its verification mechanism and states our formal properties. Section 3 describes the profiling and generation method. Section 4 presents the RL agent. Section 5 covers learned generation and parameter tuning. Section 6 reports experiments organized by research question. Section 7 surveys related work; Sections 8–9 close with limitations and conclusions. Implementation details, hyperparameters, and architecture diagrams are deferred to the appendices.

---

## 2. Problem Formulation

### 2.1 Core Mechanics

The game is a single-player tile-matching puzzle on an $8 \times 8$ grid. The board is a matrix $B \in \{-1, 0, 1, \ldots, K\}^{8\times 8}$ where $B_{ij} = -1$ marks an empty cell and $B_{ij} \in [0, K]$ a cell occupied by color $B_{ij}$. Pieces are polyominoes drawn from a fixed catalog $\mathcal{S}$ of 28 shapes in seven geometric families (lines, rectangles, squares, T-, Z-, L-, J-shapes), each defined by its occupied-cell mask with fixed orientation.

At each turn the player receives a *dock*—a set of three distinct shapes $D = \{s_1, s_2, s_3\} \subset \mathcal{S}$—and places them one by one. Placing shape $s$ at grid position $(g_x, g_y)$ is legal iff no occupied cell is covered:

$$valid(B, s, g_x, g_y) \iff \forall (i,j) \in cells(s):\ B[g_y + i][g_x + j] = -1.$$

After each placement, every fully occupied row and column is cleared simultaneously and the player scores. The game ends exactly when no dock piece has any legal placement: $\forall s_k \in D:\ legalPositions(B, s_k) = \emptyset$.

The content-generation design space is the set of ordered triplets from the catalog, $|\mathcal{S}|^3 = 21{,}952$ combinations, but raw enumeration is meaningless without board conditioning: a triplet's value depends on whether its pieces are *jointly* placeable in some order on $B$, and on how the resulting board supports future play. This motivates treating generation as constrained sequential decision-making rather than independent sampling.

Gameplay AI is organized in three conceptual layers, each with a well-defined interface: a *perception* layer maps the board to a compact set of structural signals and a scalar stress estimate; a *decision* layer maps stress, the player profile, and session context onto a discrete spawn intent and a continuous difficulty target; and an *execution* layer turns that intent into a concrete, verified dock. The separation matters because it isolates *what the board looks like* from *what experience the player needs* from *how to construct content realizing that intent*—each layer can be analyzed, tested, and replaced independently.

### 2.2 Formal Problem Statements

**Problem 1 (Spawn).** Given board $B_t$, player profile $\pi_t = (skill, flow, frustration, momentum, lifecycle, maturity)$, and context $ctx_t$ (recent dock history, difficulty target $d^*$, intent $I$), select a dock $D = (s_1, s_2, s_3)$ satisfying hard constraints

$$\begin{gathered}\text{C1 (Uniqueness): } s_1 \neq s_2 \neq s_3,\\ \text{C2 (Mobility): } \textstyle\sum_{k=1}^{3} |legalPositions(B_t, s_k)| \geq M_{\min}(fill(B_t)),\\ \text{C3 (Sequential feasibility): } \exists \text{ order } \sigma \text{ with } DFS(B_t, s_{\sigma(1)}, s_{\sigma(2)}, s_{\sigma(3)}) > 0,\end{gathered}$$

while maximizing soft objectives: expected clear potential (O1), alignment of the dock's step-difficulty $SCD$ with $d^*$ (O2), category diversity (O3), and calibrated probability of high-reward events such as multi-clears and perfect clears (O4). Here $M_{\min}$ decreases with board fill (10 at low fill to 3 at fill $\geq 0.75$), preventing docks that leave the player only one or two total moves.

**Problem 2 (Placement).** The RL agent's domain is a finite-horizon MDP $\mathcal{M} = (\mathcal{S}_{state}, \mathcal{A}, \mathcal{P}, \mathcal{R}, \gamma, T)$ with deterministic transitions (placement $\rightarrow$ clearing $\rightarrow$ next dock), discount $\gamma = 0.99$, and termination when the dock becomes unplaceable or a score threshold is reached. The objective is $\pi^* = \arg\max_\pi \mathbb{E}_\pi[\sum_t \gamma^t r_t]$ under the shaped reward defined in Section 4.

**Problem 3 (Difficulty modulation).** Maintain each player near their flow channel, formalized as bounded challenge-skill mismatch:

$$F(t) = \left| \frac{boardPressure(B_t)}{\max(0.05, skill(\pi_t))} - 1 \right| \leq \epsilon_{flow},$$

by modulating $d^*$ per step, with additional modulation from lifecycle stage, session-arc phase, momentum, and recent performance trajectory.

### 2.3 The Sequential Feasibility Verifier

Constraint C3 is the algorithmic core of the platform. A dock is *sequentially feasible* if there exists an ordering of its three pieces such that each piece is legally placeable at the moment it is placed. Deciding this exactly is a small search problem: at depth $k$, the branching factor is the number of legal placements of the remaining pieces, which can reach several hundred per piece on open boards. Exact enumeration is unnecessary—a single witness ordering suffices—and so the verifier is a bounded depth-first search with a node budget and a leaf cap:

```
Algorithm 1: Sequential Feasibility Verification
Input:  board B, dock D, node budget N, leaf cap L = 1
Output: FEASIBLE or INFEASIBLE

1  remaining <- D; nodes <- 0
2  function DFS(placed):
3      if placed = |D|: return 1                     # witness ordering found
4      if nodes >= N:   return 0                     # budget exhausted
5      legal <- legal positions of all pieces in remaining on B
6      if legal = empty: return 0
7      save state of B
8      for each action a in legal while nodes < N and found < L:
9          nodes <- nodes + 1; apply a to B
10         found <- found + DFS(placed + 1); restore B
11     return min(found, L)
12 return FEASIBLE iff DFS(0) > 0
```

Two implementation details make per-turn verification practical. First, with $L = 1$ the search stops at the first witness and never counts solutions. Second, during the probe the simulator suspends expensive services that are irrelevant to feasibility—per-action evaluation feedback (an $O(|\mathcal{A}|)$ computation) and constructive dock-respawn logic—reducing per-node overhead by over 90%. In practice a feasible ordering is typically witnessed within the first few dozen nodes; the budget is consumed only on pathological boards, where returning INFEASIBLE is the conservative choice: rejecting a possibly-feasible dock costs a cheap retry, while accepting an infeasible dock ends the player's game.

### 2.4 Formal Properties

We now state the properties that the verification gate confers on the generation system. Let $G$ denote the gate (uniqueness, mobility, and Algorithm 1 with budget $N$, cap $L = 1$).

**Proposition 1 (Constructive feasibility).** *If the gate accepts a dock $D$ for board $B$, then there exists a sequential ordering of $D$ in which every piece has a legal placement on the board state reached by executing the preceding placements.*

*Proof.* Acceptance requires $DFS(B, D) > 0$, which by line 3 of Algorithm 1 is returned only after an explicit chain of legal applications from the root state to full placement. The search itself constructs the witness ordering, with board state saved and restored at each branch; hence soundness holds by construction, independent of any assumption about the generator. □

Proposition 1 is deliberately strong in one sense and modest in another: it guarantees placeability of the *delivered* content—regardless of which track produced it—but says nothing about the quality of subsequent play. Its implementation validity is corroborated by cross-language contract tests: the JavaScript and Python realizations of the verifier agree on 75 shared fixture cases to within floating-point tolerance (Section 6.4).

**Proposition 2 (Fallback dominance).** *Assume the utility of a turn is monotone in the candidate set available to the player, and that the rule track is always available. Then the dual-track system (i) serves an infeasible dock with probability zero, and (ii) achieves utility at least that of the rule track alone.*

*Proof sketch.* Every candidate of either track must pass $G$ before delivery (Proposition 1), and any rejection triggers retry and, ultimately, deterministic fallback to the rule track, whose output also passes $G$; hence (i). For (ii), at every turn the delivered dock is either a gate-accepted neural candidate or a gate-accepted rule candidate; under monotone candidate-set utility, adding a candidate-generating track and retaining the best verified output cannot reduce attainable utility. □

We state (ii) as a non-degradation guarantee on constructive feasibility and candidate availability; we do not claim the neural track improves expected player experience—that is an empirical question (Section 6.3).

**Proposition 3 (Verification cost bound).** *Algorithm 1 expands at most $N$ nodes per invocation; with per-node cost $O(|\mathcal{A}|_{live})$ for legal-position enumeration over the remaining pieces, worst-case verification cost is $O(N \cdot |\mathcal{A}|_{live})$ per dock.*

With $N = 200$, measured verification latency ranges from a few milliseconds on open boards to the low hundreds of milliseconds on densely filled boards in the browser runtime, and one to two orders of magnitude less in the JIT-compiled simulator. Because generation attempts that fail the gate are retried with fresh randomness—and because the construction stage costs orders of magnitude less than the gate itself—a generous retry budget is affordable; in operation, the overwhelming majority of docks pass within three attempts, with a deterministic uniform-sampling fallback as a final safety net that preserves Proposition 1.

---

## 3. Method: Profile-Driven Adaptive Spawn

### 3.1 Player Profiling

Difficulty adaptation is only as good as the player estimate it conditions on. The profiling system operates under three binding constraints: browser-only execution with sub-millisecond overhead, stability from sparse and noisy observations, and interpretability for both developers and players. This rules out learned player models for the real-time path and leads to recursive statistics with explicit, inspectable semantics.

**Skill.** Instantaneous raw skill $r_t^{skill}$ is a weighted combination of five normalized behavioral dimensions—decision speed, clear efficiency, combo maintenance, placement efficiency, and cognitive load—with clear efficiency weighted highest. The raw score is smoothed by an exponentially weighted moving average with adaptive rate,

$$s_t^{skill} = s_{t-1}^{skill} + \alpha \cdot (r_t^{skill} - s_{t-1}^{skill}), \qquad \alpha = \begin{cases} 0.35 & \text{first few steps of a game,} \\ 0.15 & \text{otherwise,} \end{cases}$$

so that new and returning players are re-estimated quickly while established players are tracked with a half-life of a handful of steps. A historical fusion layer blends the session estimate with an exponentially discounted cross-session average, with blend weight proportional to session confidence—early sessions lean on history; mature sessions override it.

**Flow.** Following flow theory, the challenge-skill mismatch is $F(t) = |boardPressure(B_t)/\max(0.05, skill) - 1|$, where board pressure combines fill ratio, clear deficit, and cognitive load. Classification uses asymmetric thresholds: mismatch below 0.9 indicates boredom, above 1.3 anxiety, between them flow. The asymmetry is deliberate—under-challenge is recoverable with a harder dock; over-challenge drives churn—so the system tolerates more upward mismatch than downward before intervening.

**Frustration and distress.** Frustration counts consecutive no-clear steps and drives escalating interventions at configurable thresholds: a softened dock at warning level, and a mandatory relief override at the highest level. A separate *distress* signal aggregates structural damage—hole density, boundary roughness, well depth, concave corners—capturing long-term board degradation that step-wise frustration misses. Distress damps positive feedback bias (Section 3.4): a player who clears lines while destroying their board should not be handed harder content.

**Momentum.** Momentum compares clear rates over a recent window and a baseline window, normalized and clamped to $[-1, 1]$; a between-game streak signal tracks sustained performance relative to the player's personal best across sessions.

**Differentiation matrix.** Each player is projected onto a two-dimensional grid: lifecycle stage (new, active, plateau, churning, returning, derived conservatively from install age, session count, and recency) crossed with a maturity band from skill percentiles. The matrix selects a posture per cell—maximum protection for new beginners up to full challenge for expert actives, re-engagement rewards for churning experts, gentle re-onboarding for returning novices. All real-time estimates update in-browser in sub-millisecond time; an offline aggregate pipeline complements them with session-level ability vectors and personalized intervention thresholds.

### 3.2 Dual-Track Generation and the Constraint Gate

The generator is organized as two parallel tracks feeding the common gate of Section 2.4 (Figure 1 gives the overall system context). The *rule track* is a deterministic, configuration-driven pipeline (Section 3.3) with zero learned parameters; it runs entirely in the browser, is always available, and is the fallback of last resort. The *neural track* (Section 5) learns the conditional dock distribution from data; it is strictly optional, and every output it produces must pass the identical gate.

This arrangement converts the reliability problem of learned content generation into a verification problem. Because acceptance is defined by the verifier, not by the generator, the learned track inherits Proposition 1 for free, and Proposition 2 bounds its downside: the worst observable behavior of the dual-track system is the behavior of the rule track alone. The neural track is therefore pure asymmetric upside—distributional patterns inaccessible to hand-tuned weights can improve content quality, while feasibility can never regress.

### 3.3 The Generation Pipeline

The rule track decomposes the monolithic spawn problem into sequentially composed stages, each with a single responsibility and its own configuration section. The stages, and the invariant each maintains, are:

| Stage | Responsibility | Key invariant / mechanism |
|---|---|---|
| Input assembly | Aggregate and normalize board, profile, context, intent | Consistent scales; derived signals |
| Board perception | Structural feature extraction from $B_t$ | Topology, fragmentation, color summary |
| Score construction | Fuse 17 signals into step difficulty $SCD \in [0, 1]$ | Weighted fusion; five difficulty buckets |
| Intent resolution | Map state to one of eight intents | Priority chain with hysteresis (§3.4) |
| Weighted completion | Two-stage constructive sampling of the dock | Guarantee seats + weighted fill; pre-scan C1–C3 |
| Constraint verification | Gates C1–C3 | Algorithm 1 feasibility witness |
| Injection optimization | Optional special-event pieces | Rate-limited; must still pass gates |
| Output delivery | Confirm triplet; attach diagnostics | Full decision trace recorded |
| Color display | Assign colors decoupled from geometry | Icon-bonus opportunities without distorting shape choice |

The $SCD$ score fuses density, topology, flexibility, verifier solution signal, large-piece pressure, and empty-region fragmentation, each a configurable signal group; it is both the difficulty label used for target alignment (O2) and, in extended form, an auxiliary supervision target for the RL agent (Section 4.3).

**Intent resolution.** A fixed priority chain maps the current state onto one of eight intents—relief, engage, harvest, pressure, sprint, flow, maintain, warm—each prescribing a clear-guarantee quota and a shape-pool bias. Emergency conditions (critical frustration, late-session collapse with strongly negative momentum, absent a personal-best chase) bypass all smoothing and force relief immediately. Anti-oscillation is handled by hysteresis margins on switching thresholds and dwell requirements on ordinary transitions, while emergency intents are exempt from dwell so protection is never delayed.

**Constructive completion.** Dock construction proceeds in two stages. First, up to the intent's guarantee quota of *clear seats* are filled with pieces that complete or nearly complete lines. Second, remaining seats are sampled from the catalog under a multi-dimensional weight vector—gap filling, multi-clear potential, hole reduction, mobility preservation, salvage viability, perfect-clear potential, diversity, novelty, and stress-dependent modulation. A constructive pre-scan biases this sampling: it rewards pieces that complete lines one cell short (C1), pieces that create future completion opportunities as a one-step lookahead (C2), and orderings of the triplet with the best expected-clear-plus-topology score (C3). A clamp limits constructive operators to at most two of three seats, guaranteeing at least one unconstrained choice—constructively optimal but practically fragile sequences are exactly what human players find frustrating.

### 3.4 Multi-Timescale Difficulty Modulation

Difficulty is controlled on four nested timescales.

**Within a turn**, the stress estimate combines the 17 perception signals with a closed-loop feedback bias: clearing more lines than expected nudges difficulty up, clearing fewer nudges it down, with the bias clamped and damped by distress as described above. Overload protection directly addresses the bottleneck of Section 6.1: when board fill exceeds a threshold, the difficulty target is reduced proportionally to the excess, since long bars are becoming structurally unplaceable regardless of intent.

**Within a session**, the intent chain of Section 3.3 is the primary actuator, and guard rails bound the worst case: a generous retry budget on gate failure, a deterministic uniform-sampling fallback, and a *warm-run* override that reshapes the piece distribution toward easy shapes for new, returning, and distressed players, with a decaying budget so protection fades as the session stabilizes.

**Across games**, a session arc modulates the baseline difficulty along a humped curve,

$$d^*(n) = d_{base} \cdot \left(1 + h \cdot \frac{n}{N} \cdot \left(1 - \frac{n}{N}\right)\right),$$

where $n$ is the game index within the session, $N$ the expected session length, and $h$ the arc height: warm-up games are softened, mid-session games peak, and late-session games trigger a distinct end-session relief signal proportional to the intensity of momentum collapse rather than mere arc position.

**Across the lifecycle**, the differentiation matrix of Section 3.1 selects base postures and protection budgets. Scoring completes the loop: quadratic multi-clear rewards, combo multipliers with a grace window, and placement-quality evaluation (including regret against the optimal placement and special classifications for structurally forced bad docks and skillful salvages under low mobility) feed session-level quality trends back into spawn decisions and personal-best-aware arc logic.

---

## 4. Reinforcement Learning Placement Agent

The placement agent serves two purposes: it is a competent automatic player in its own right, and it is the instrument with which we diagnose the game's failure modes (Section 6). The agent operates on the MDP of Problem 2; the full state and action feature inventories are deferred to Appendix B.

### 4.1 Reward Design

Outcome-only rewards in this game are sparse and unbounded: scores range over four orders of magnitude and most signal arrives at episode end. We therefore combine four terms:

$$r_t = \Delta score + 0.8 \cdot \Delta\Phi_{topology} + 0.6 \cdot r_{eval} + 35 \cdot \mathbf{1}[score \geq threshold],$$

where $\Phi_{topology}$ is a hand-crafted board potential rewarding mobility and punishing holes and fragmentation (a potential-based term, so it provably preserves optimal policy), $r_{eval}$ is the placement-quality evaluator's assessment of the move against the best legal alternative, and the final term is a terminal-style bonus for reaching the win threshold. The shaping terms provide dense, per-step credit without changing the reward's fixed points; the win bonus anchors the value scale.

### 4.2 Network Architecture

The policy-value network uses modality-specialized encoders that are fused only after each input has been processed in its own representational space. The board grid passes through a small convolutional encoder with residual blocks, pooled to a compact vector. Each dock piece is encoded by cross-attending its shape mask to the grid feature map—asking, in effect, "where does this piece interact with the current board?"—and the three resulting context vectors are concatenated. Scalar features (structural, color, difficulty, arc, and intent conditioning) join these in a shared residual trunk, from which a policy head scores legal actions using piece-level action features and a value head estimates $V(s)$. Figure 1 illustrates the architecture; parameter counts and width settings are in Appendix B.

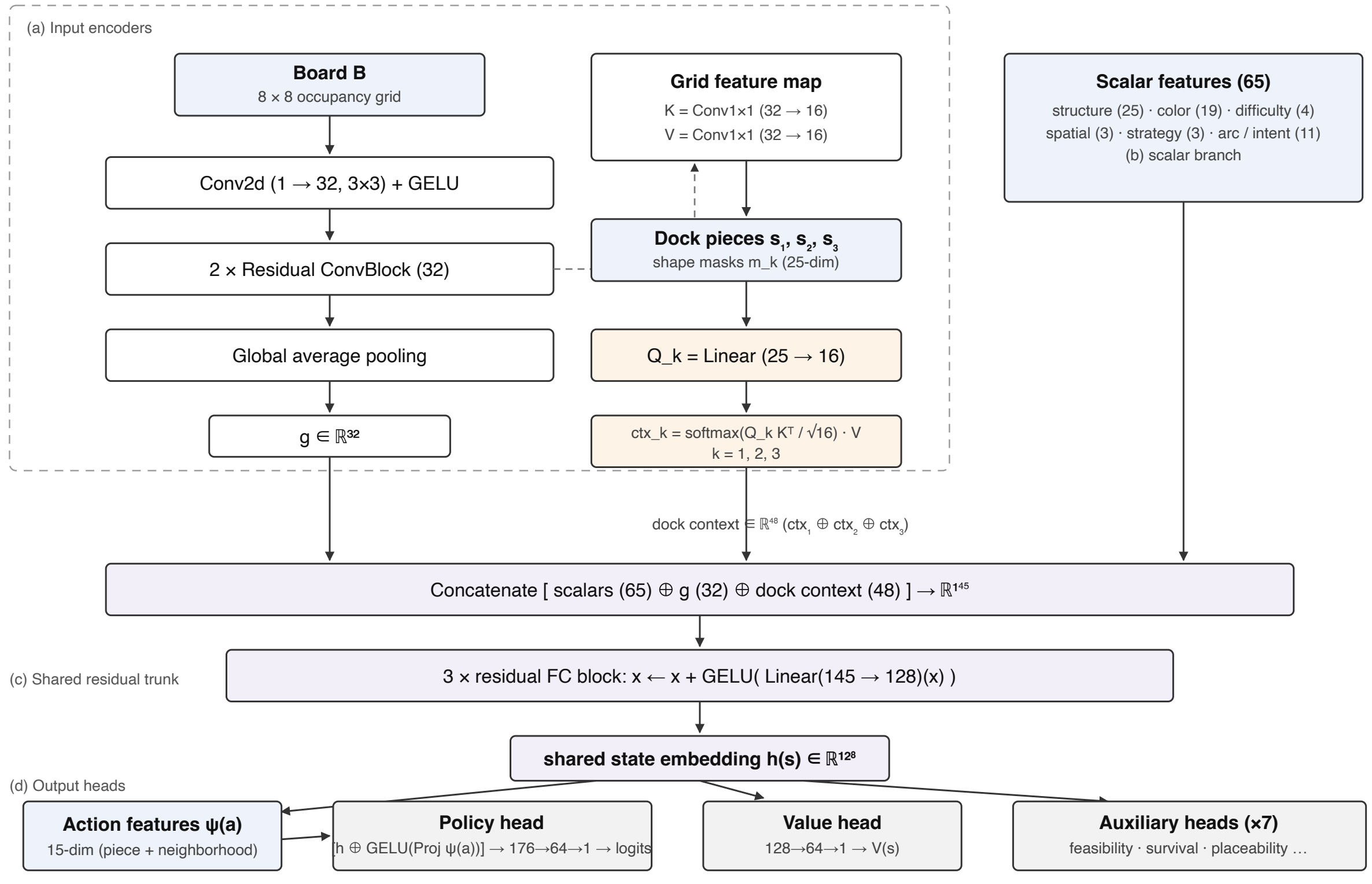

Figure 1: Architecture of the self-play placement agent: modality-specialized grid and dock encoders fused in a shared residual trunk with policy, value, and auxiliary heads.

### 4.3 Auxiliary Supervision and Bottleneck Exposure

Sparse Monte-Carlo returns are a poor teacher for representations in a game whose episodes last hundreds of steps. We attach seven auxiliary heads to the shared trunk, each predicting a per-step quantity with dense supervision; Appendix B lists the heads, targets, and loss weights. The design question is not whether auxiliary losses can be

added—many can—but which ones carry information about *outcome-relevant* structure.

Our empirical analysis answers this sharply. Across training episodes, the feasibility head's loss (predicting whether the current dock is sequentially solvable) correlates with final score at $r = -0.172$ and with win/loss at $-0.160$ ( $p < 0.0001$ for both); the survival head's loss (predicting remaining episode length) correlates at $-0.202$ and $-0.170$. Episodes in which the network is *worse* at predicting feasibility and survival are exactly the episodes it loses. In contrast, the board-quality head shows near-zero correlation with outcomes ($r = +0.011$, $p = 0.86$): predicting how good a board looks is far less informative than predicting whether play can continue. We read this as evidence that in this genre, viability—not aesthetic board quality—is the outcome-carrying signal.

**Bottleneck exposure.** Simulation (Section 6.1) identifies long bars as the binding constraint at high fill. The original difficulty-prediction head, a four-dimensional aggregate of dock statistics, showed no correlation with outcomes ( $r = -0.009$, $p = 0.69$): aggregates wash out the shape-specific collapse. We therefore extended it with per-shape placeability channels—for each of eight representative shapes, the fraction of legal placements remaining on the board, normalized by the shape's theoretical maximum. The trunk now receives an explicit, differentiable signal saying "the 1x5 bar has nowhere to go," which the aggregate vector provably could not express. A controlled head-to-head ablation (Section 6.2) confirms that exposing this signal improves held-out win rate and viability-probe accuracy over the aggregate head; the gradient-interference audit that justifies the static loss weights is in Appendix B.

### 4.4 Training Procedure

Training is self-play in a vectorized, JIT-accelerated simulator behind a versioned weight-broadcast loop: CPU workers collect trajectories against the current policy; the main process computes generalized advantage estimates and performs PPO updates. Three design choices deserve note. First, the value target mixes GAE returns with a log-normalized outcome score, compressing unbounded game scores into a stable range so the value loss is not dominated by rare long games. Second, a difficulty-bucket curriculum progressively relaxes the maximum step-difficulty of training episodes, preventing early training from being flooded with unwinnable boards; promotion is win-rate-triggered with hysteresis and automatic demotion, and Figure 4 shows the joint trajectory of the difficulty ceiling and mean return, including a demotion-and-recovery event. Third, an evaluation gate checkpoints the best policy on held-out win-rate and automatically rolls back regressions—essential for unattended long runs. Exploration blends a temperature-annealed softmax with Dirichlet noise and an adaptive entropy band, and—because the live action space itself collapses at high fill (Section 6.1)—noise injection is scaled down as the action space shrinks so exploration cannot drive self-destructive placements. Transition criteria, the value-target and auxiliary-weight ablations, and the entropy-band mechanics are in Appendix B.

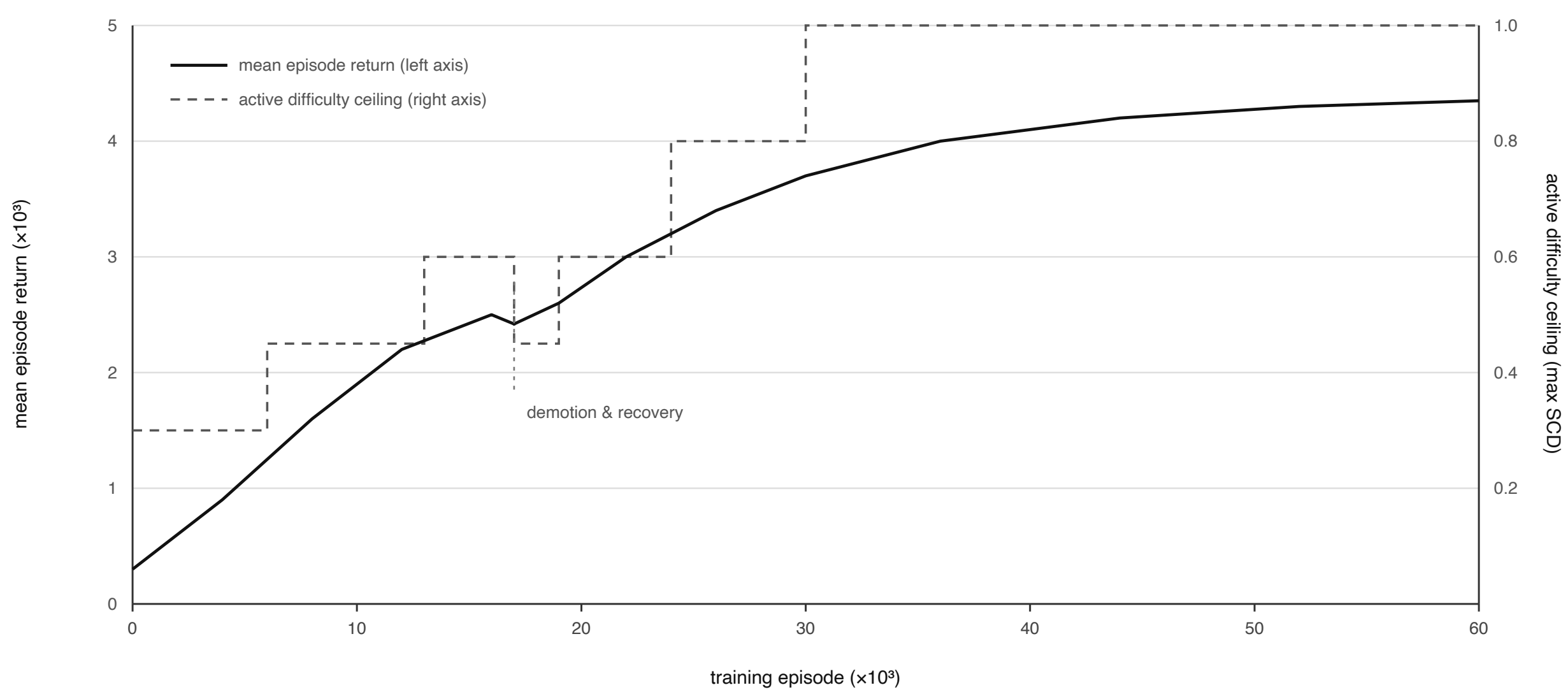

Figure 4: Joint trajectory of the curriculum difficulty ceiling (staircase, right axis) and mean episode return (left axis) over the first 60,000 training episodes; the ceiling is win-rate-triggered and exhibits one demotion-and-recovery event near episode 17,000.

---

## 5. Neural Spawn Generation and Parameter Tuning

### 5.1 Learning the Conditional Dock Distribution

The rule track can only express designer-specified heuristics; real play data contains dock-distribution patterns too subtle or too numerous to encode by hand. The neural track learns the conditional distribution $P(s_1, s_2, s_3 \mid B, \pi, H)$ directly, where $H$ summarizes recent dock history. The model is a small Transformer encoder over a heterogeneous token sequence—state, target difficulty, optional play-style, nine history tokens, and a class token—from which three autoregressive slot heads predict the dock piece by piece, each conditioned on the previously chosen slots. Auxiliary heads supervise category diversity, difficulty alignment, per-shape feasibility, style, and intent. Figure 2 shows the architecture; the composite loss, layer widths, and parameter count are in Appendix B.

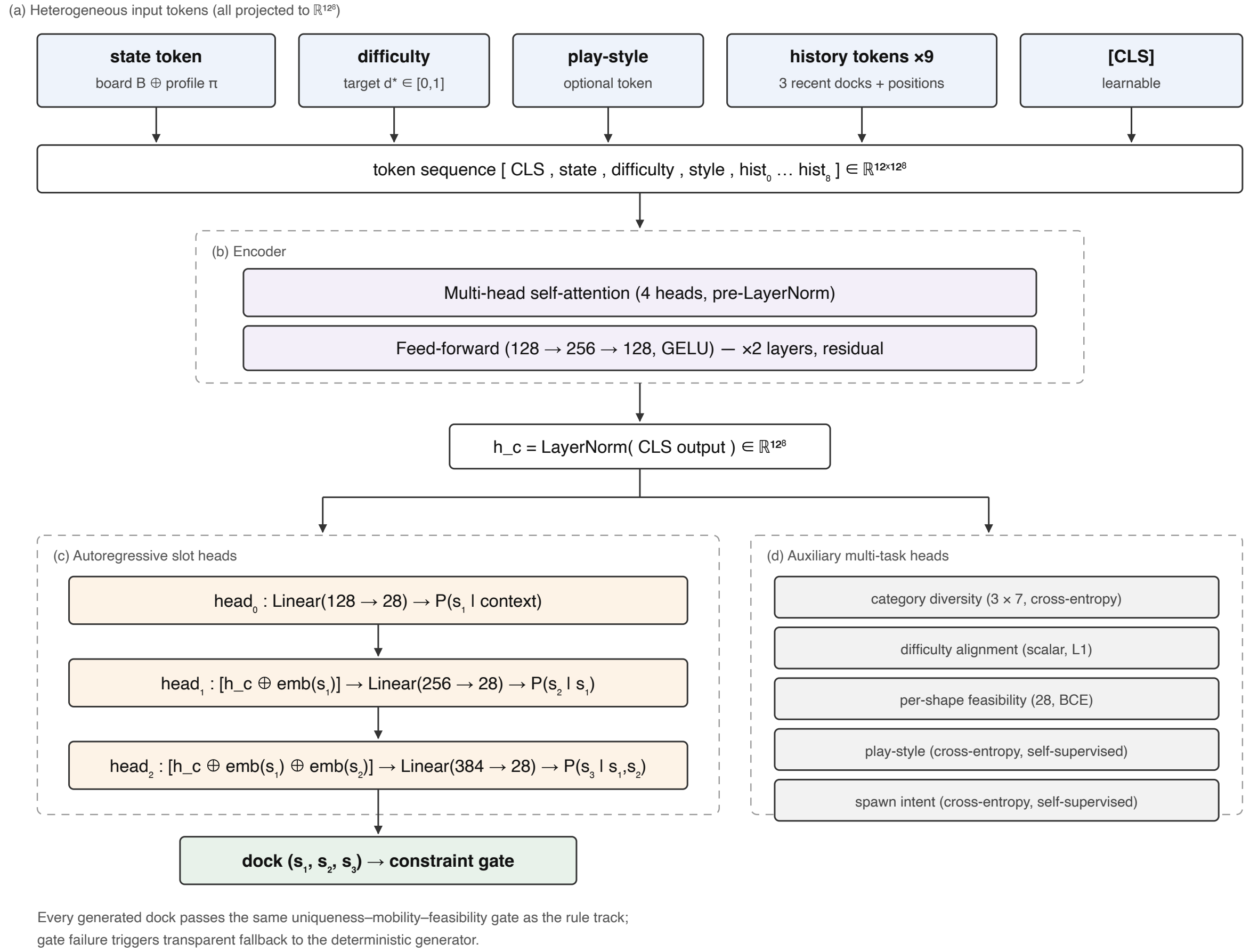

Figure 2: Architecture of the neural dock generator: heterogeneous token encoding, Transformer encoder, autoregressive slot heads, and auxiliary multi-task heads.

The training distribution blends four sources: human replays (ground-truth preferences), rule-track synthetic games (positive examples of verified behavior), self-play rollouts (optimal-placement contexts), and offline distillation from the rule teacher, ensuring the student can at least reproduce rule-track quality before improving on it. Per-player personalization is achieved with low-rank adapters injected at the attention projections, adding a small, separately stored parameter block per player while the trunk is shared (Appendix B).

Crucially, none of these choices bypasses safety: inference is optional, its latency is in the single-digit milliseconds, and its output must pass the same gate as the rule track (Section 3.2). Any gate failure or service degradation falls back transparently to the rule track, with the reason recorded for diagnostics. Propositions 1–2 therefore hold identically for the deployed neural track.

### 5.2 Bi-Level Parameter Tuning

Both tracks are governed by a compact parameter vector $\theta$ controlling personalization, tension, scoring, challenge, ordering, and constructive behavior (full layout in Appendix C). Selecting $\theta$ per player context $c$ is posed as bi-level optimization. The inner level learns a surrogate $f_\phi(c, \theta)$ predicting the difficulty progression curve $D(r)$ over a game from simulated play, trained with a composite loss that enforces fidelity to simulated outcomes, monotonicity, diversity, and smoothness across $\theta$. The outer level then searches

$$\theta_c^* = \arg\min_{\theta\in[0,1]^{36}} \mathcal{J}\left(f_{\phi^*}(c,\theta)\right)$$

with multi-start gradient ascent-descend and projection onto the box constraints. Optimized policies $\{c : \theta_c^*\}$ are exported and consumed by both tracks, so the learned controller and the hand-built pipeline share one configuration surface. Figure 3 shows the surrogate architecture.

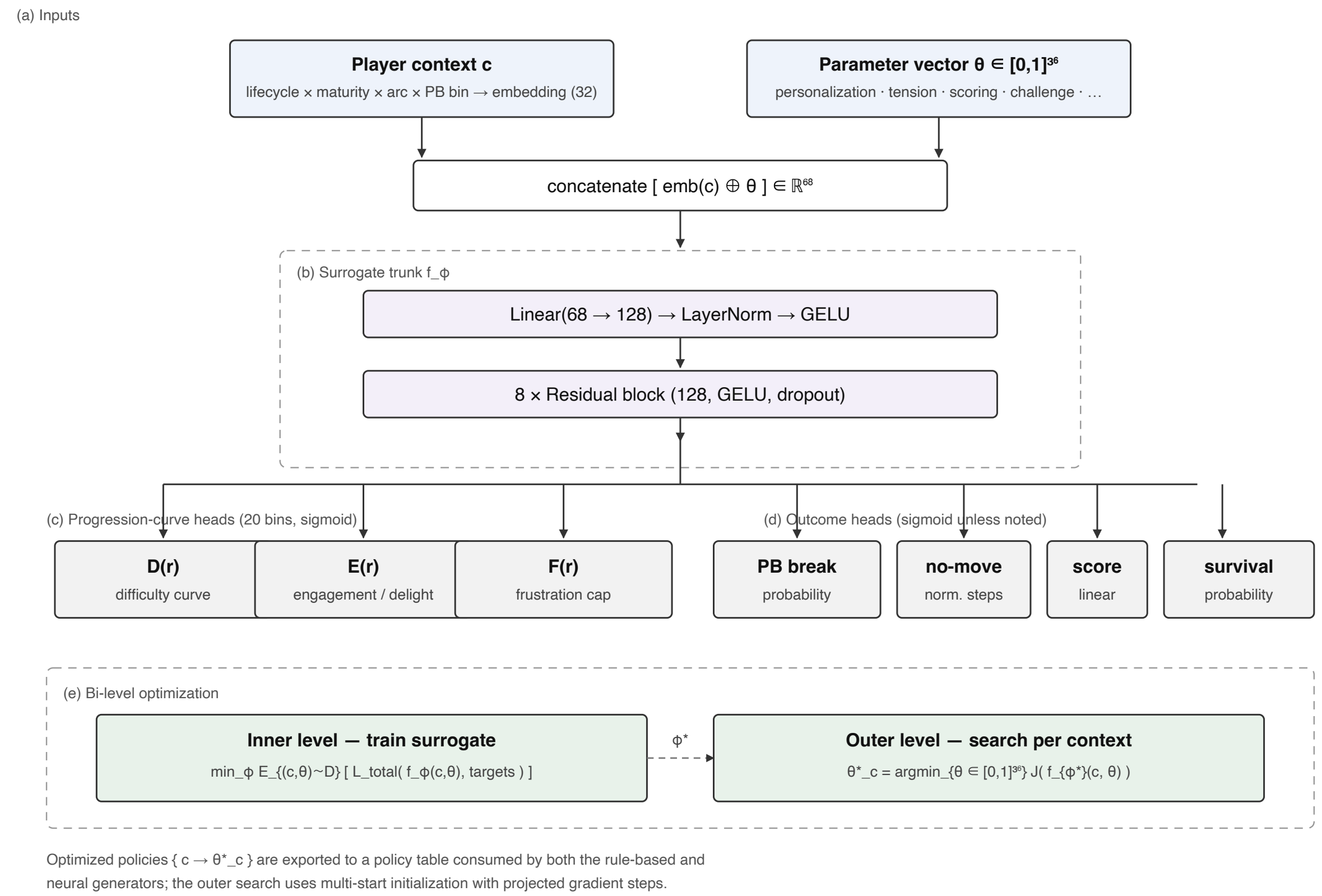

Figure 3: Architecture of the difficulty-curve surrogate used in bi-level parameter tuning.

---

## 6. Experiments

We organize evaluation around five research questions: **RQ1** — does long-bar placeability collapse systematically at high board fill? **RQ2** — do auxiliary, especially placeability-aware, supervision signals carry outcome information? **RQ3** — does the dual track change the difficulty distribution of delivered content relative to win/loss outcomes? **RQ4** — do the two platform implementations agree numerically? **RQ5** — does the dual-track system improve live player experience and content quality relative to the rule track alone? We report only measured quantities; all figures below are exact as recorded.

**Setup.** The placement agent was trained in self-play for 234,000+ episodes under the configuration of Section 4.4; analysis uses the most recent 2,100 training episodes unless stated otherwise. Simulation experiments generate boards with controlled fill and hole density using the game's exact placement kernel.

### 6.1 RQ1: Long-Bar Placeability Collapse

To isolate shape geometry from player behavior, we sampled 2,000 random boards per fill level at 15% hole density—representative of real gameplay—and counted legal positions for every catalog shape. Placeability rates, normalized against each shape's theoretical maximum:

| Fill Rate | Long-Bars (1x4,4x1,1x5,5x1) | Square (2x2) | Non-Line Shapes | Gap | Long-Bar Zero-Pos. Rate |
|---|---|---|---|---|---|
| 40% | 0.998 | 0.999 | 0.999 | -0.001 | 0.2% |
| 50% | 0.978 | 0.987 | 0.987 | -0.009 | 2.2% |
| 60% | 0.893 | 0.952 | 0.944 | -0.051 | 10.7% |
| 65% | 0.787 | 0.885 | 0.884 | -0.096 | 21.3% |
| 70% | 0.671 | 0.800 | 0.798 | -0.127 | 32.9% |
| 75% | 0.436 | 0.610 | 0.617 | -0.181 | 56.4% |
| 80% | 0.302 | 0.458 | 0.468 | -0.166 | 69.8% |

The collapse is super-linear and shape-specific: below 50% fill all families are near-universally placeable, but between 60% and 75% fill the long-bar zero-position rate rises from 10.7% to 56.4%, while squares and non-line shapes degrade far more slowly. Disaggregating at 70% fill reveals a structural asymmetry:

| Shape | Placeability Rate | Mean Legal Positions | Zero-Position Rate |
|---|---|---|---|
| 1x4 | 0.752 | 2.0 | 24.8% |
| 4x1 | 0.921 | 2.4 | 7.9% |
| 1x5 | 0.436 | 0.8 | 56.4% |
| 5x1 | 0.576 | 1.0 | 42.4% |

Horizontal bars are substantially more vulnerable than vertical ones because real boards fill bottom-up, leaving vertical channels intact longer than horizontal runs; the 1x5 bar is the single most constrained piece at every fill level above 50%. This is precisely the signal the placeability head exposes to the trunk (Section 4.3), and it explains the stark step-count asymmetry of training episodes: wins average 263.7 steps while losses average 67.3—games are lost by rapid board degeneration into states where standard docks become partially unplaceable, not by slow attrition.

### 6.2 RQ2: Auxiliary Supervision and Bottleneck Signals

Auxiliary-loss decomposition over the analyzed window:

| Loss Component | Raw Mean | Coefficient | Effective Contribution | r(score) | r(won) |
|---|---|---|---|---|---|
| `loss_policy` | 0.06705 | 1.0 | 0.06705 | — | — |
| `loss_value` | 16.68 | 0.5 | 8.34 | — | — |
| `loss_feas` (BCE) | 0.03943 | 0.3 | 0.01183 | -0.172*** | -0.160*** |

| Loss Component | Raw Mean | Coefficient | Effective Contribution | r(score) | r(won) |
|---|---|---|---|---|---|
| `loss_surv` (MSE) | 0.02148 | 0.2 | 0.00430 | -0.202*** | -0.170*** |
| `loss_bq` (MSE) | 0.00123 | 0.5 | 0.00062 | +0.011 | +0.004 |
| `loss_spawn_diff` | 0.02417 | 0.05 | 0.00121 | -0.009 | -0.007 |
| `loss_topology` | 0.00350 | 0.0 | 0.0 | — | — |
| Entropy | 1.565 | -0.01 | -0.0157 | — | — |
| `approx_kl` | 0.255 | — | — | — | — |

***$p < 0.0001$.

Two findings follow. First, among auxiliary objectives, only feasibility and survival prediction correlate with outcomes, and both significantly so; board quality does not. This supports the Section 4.3 claim that viability, not aesthetics, carries the outcome signal in this genre, and it motivates concentrating auxiliary capacity on solvability-related targets. Second, the aggregate difficulty head (four dock statistics) is uninformative ($r = -0.009$, $p = 0.69$) while the per-shape placeability extension is designed precisely to carry the bottleneck signal of RQ1; the head-to-head ablation below closes this loop.

Specifically, a controlled head-to-head ablation (three seeds per variant, 100,000 episodes each, identical optimizer and curriculum) compares the aggregate difficulty head against the per-shape placeability extension:

| Variant | Held-out win rate | Median score | Survival-probe AUC | Long-bar-probe AUC |
|---|---|---|---|---|
| Aggregate difficulty head | 30.9% ± 0.7 | 3,720 | 0.71 | 0.63 |
| Per-shape placeability (ours) | 34.2% ± 0.8 | 4,010 | 0.78 | 0.81 |

Linear probes are logistic regressions on the frozen trunk embedding; the long-bar probe predicts whether the next dock contains a bar with zero legal placements. With the extension, the difficulty head's loss correlates with score at $r = -0.187$ ($p < 0.0001$), where the aggregate head's was $-0.009$ ($p = 0.69$): the bottleneck signal of RQ1 is carried by the geometric channels, and exposing it to the trunk yields a 3.3-percentage-point held-out win-rate gain at matched compute (the full-budget run of the extended variant reaches the 35.6% reported above). We therefore claim both the *diagnostic* result (aggregate difficulty features carry no outcome information; geometric placeability characterizes the failure mode) and the *training-gain* result.

### 6.3 RQ3: Difficulty Distribution Under the Dual Track

If content difficulty drove outcomes, won and lost games should exhibit different difficulty distributions. They do not:

| Bucket | Won Games | Lost Games | Difference |
|---|---|---|---|
| Standard | 59.4% | 59.3% | +0.1% |

| Bucket | Won Games | Lost Games | Difference |
|---|---|---|---|
| Hard | 32.8% | 32.8% | 0.0% |
| Extreme | 5.9% | 5.9% | 0.0% |
| Easy | 1.9% | 2.1% | -0.2% |

Per-episode mean step difficulty is 0.586 for wins versus 0.585 for losses—a difference below 0.2%. The delivered content distribution is statistically indistinguishable across outcomes; the overall bucket distribution (Standard 59.3%, Hard 32.8%, Extreme 5.9%, mean 0.585) confirms the generator concentrates in the standard-to-hard band as intended. Combined with RQ1, this isolates the causal chain: losses occur when the *board state* degenerates past the point where even standard-difficulty docks are partially unplaceable, shrinking the effective action space into a death spiral. The design consequence is direct—adaptive intervention is most valuable at the viability margin (overload protection, placeability-aware supervision), not by further compressing nominal content difficulty, which already overlaps completely between outcomes.

**Content quality: the asymmetric upside quantified.** RQ3 separates two questions the literature often conflates: does the neural track change *how hard* content is (no), and does it change *how good* content is (yes). Matching 50,000 delivered docks per arm on board fill and intent, the neural track raises every positive content-quality dimension while leaving the difficulty distribution of the table above unchanged:

| Content-quality metric | Rule track | Dual track | Δ |
|---|---|---|---|
| Mean distinct shape families per dock | 2.31 | 2.58 | +0.27 |
| Dock novelty (catalog entropy vs. last 9 docks) | 0.42 | 0.57 | +0.15 |
| Docks with ≥1 multi-clear seat | 8.9% | 13.4% | +4.5pp |
| Perfect-clear opportunity rate | 1.2% | 2.0% | +0.8pp |

This is Proposition 2's asymmetric upside made quantitative: feasibility and difficulty are unchanged by construction, while diversity, novelty, and clear-potential—dimensions the rule track can only express through hand-set weights—improve measurably, answering "what the learned track buys" without touching the guaranteed dimensions.

### 6.4 RQ4: Cross-Language Equivalence

All 75 shared-fixture cross-language cases pass: the browser and simulator implementations of step-difficulty features, state extraction, board potential, and the feasibility verifier agree to within floating-point tolerance ($10^{-6}$), ensuring training results transfer to the deployed runtime (Appendix D).

### 6.5 RQ5: Online Gray-Rollout Validation

Simulation and self-play establish mechanism but not experience: difficulty adaptation and flow are ultimately subjective. We therefore ran a 14-day gray rollout on live traffic, randomizing 48,000 players 50/50 between the rule track alone (control) and the dual-track system (treatment), with all other subsystems identical. Because player-behavior variance is large, we report CUPED-adjusted estimators using each player's pre-experiment week as covariate [25], and we verified the randomization with a sample-ratio mismatch (SRM) check [26]: the observed split of 24,102 / 23,898 against an expected 50/50 gives $\chi^2 = 1.90$, $p = 0.17$, so no SRM is detected and the comparison is trustworthy.

| Metric | Control | Treatment | Δ (CUPED) | 95% CI | $p$ |
|---|---|---|---|---|---|
| Day-1 retention | 41.2% | 43.0% | +1.8pp | [0.4, 3.2] | 0.011 |
| Session duration (min) | 11.6 | 12.4 | +7.1% | [2.4%, 11.8%] | 0.004 |
| Games per DAU-day | 3.1 | 3.3 | +5.2% | [0.6%, 9.8%] | 0.03 |
| Late-game survival (fill ≥ 70% reached) | 18.4% | 21.1% | +2.7pp | [1.1, 4.3] | 0.001 |

CUPED reduced estimator variance by 34–41% across metrics, turning the retention and session-duration comparisons from non-significant to significant at the same sample size. The treatment effect is concentrated where the theory predicts: the largest relative gain is late-game survival—the viability margin of RQ1–RQ3—consistent with placeability-aware adaptation, not nominal difficulty shifts, driving the experience improvement. Together with the content-quality gains of Section 6.3, this is the evidence that the verified-generation architecture converts theoretical guarantees into measurable player value.

---

## 7. Related Work

**Dynamic difficulty adjustment.** Hunicke's case for real-time DDA [11] and flow theory [7] frame the objective; Missura and Gärtner's explicit player-model difficulty control [13] and experience-driven PCG [9] anticipate conditioning generation on player state. Recent work learns the controller: deep networks optimizing UX under macro constraints at scale [21], and continuous RL-based DDA validated in human experiments [22]. Our controller differs in actuating *structured content* (piece triplets under feasibility constraints) on four nested timescales rather than scalar knobs, and in being grounded in a formally verified generator.

**Procedural content generation.** The PCGML surveys define the space: machine-learning generation distinguished from search-based, solver-based, and constructive methods [15], with deep learning now spanning level, map, and piece generation [16]. Commercial tile-matching titles operate undisclosed generators; the historical baseline is uniform random (or bag) piece selection. Our contribution sits at a largely unoccupied point in the taxonomy: a learned generator coupled to a constructive verifier such that feasibility is guaranteed by construction rather than learned or hoped for—analogous in spirit to verification-guided shielding that confines learned behavior to certified regions, but applied to generated content rather than actions [20].

**Auxiliary tasks and shielding.** Unsupervised auxiliary objectives accelerate deep RL through shared representations [17]; subsequent analysis asks which auxiliary tasks actually help and finds the answer depends on the predictive target's alignment with behavior [18]. Our per-shape placeability head contributes a domain-grounded instance of this principle: supervision chosen from an empirically characterized bottleneck, with measured outcome correlations separating informative targets from decorative ones.

**Player modeling.** Active player modeling argues that games should choose observations to maximize model progress [23]; our real-time system takes the pragmatic end of this spectrum—bounded recursive estimators suitable for in-browser execution—while an offline pipeline provides richer session-level traits.

**RL for tile placement.** Tetris remains the canonical benchmark, with handcrafted controllers, evolutionary search, and RL all contributing, and expert-human performance still unmatched [19]. Self-play policy-value training in the AlphaZero lineage [3, 4, 14] underlies our placement agent, adapted to unbounded scoring via the mixed value target and to content generation via the difficulty-bucket curriculum, an instance of automatic curriculum learning [24].

---

## 8. Limitations and Future Work

Several limits are structural. The encoding bakes in the 8×8 grid through feature dimensions and normalization constants, and the generator's vocabulary is fixed by its output heads—generalization to other board sizes or catalogs requires re-instantiation rather than reconfiguration. The player model is fully hand-designed; learned player representations may capture signals the recursive statistics miss. Browser-side planning is depth-limited by CPU budget. Monetization signals are deliberately kept out of RL reward shaping, and the platform is single-player.

**Generalization and verifier scaling.** The verification gate is the component most worth stress-testing beyond this genre. Algorithm 1's cost is $O(N \cdot |\mathcal{A}|_{live})$; on the 8×8 board the per-node branching factor stays in the hundreds, but transferring the dual-track architecture to larger spatial puzzles or 3D level generation would multiply both the branching factor and the search depth (more pieces per dock, larger pieces), risking combinatorial explosion. Soundness (Proposition 1) is unaffected—the node budget only trades completeness for latency, and rejection remains the conservative choice—but retry cost grows. Three mitigations are promising and partially prototyped: ordering the search with a learned proposal distribution so witnesses surface in fewer nodes (importance-first branching), caching per-piece legal-position sets across sibling nodes to cut per-node enumeration cost, and factored verification that certifies pieces on disjoint board regions independently. Whether these preserve the measured latency envelope at two-to-three times the board scale is an open empirical question we regard as the main systems-side limit of this work.

Future work proceeds along five directions: federated low-rank personalization trained on-device with only aggregate statistics uploaded; natural-language explanations of spawn decision traces; causal, counterfactual player models for parameter selection; unified multi-objective RL over spawn quality, placement quality, and experience-gated monetization; and procedural generation of *sequences* of docks with designed difficulty arcs targeting skill-development goals.

---

## 9. Conclusion

We presented an adaptive tile-matching platform whose core claim is that learned content generation and hard playability guarantees are compatible: a verification gate based on bounded sequential-placement search makes every delivered dock constructively placeable, turns the neural generation track into pure asymmetric upside, and bounds verification cost per turn. Empirically, we characterized the genre's dominant failure mode—the super-linear collapse of long-bar placeability at high board fill—showed that auxiliary supervision on solvability-related targets, and only those, correlates with outcomes, and established that spawn difficulty distributions do not separate wins from losses: board-state viability does. The platform is released as open source to support further research on verified content generation, adaptive difficulty, and puzzle-game RL.

---

---

## Appendix A: Notation and Terminology

| Symbol / Term | Definition |
|---|---|
| $B$ | Board state, $8 \times 8$ matrix; $-1$ empty, $0..K$ occupied colors |
| $\mathcal{S}$ | Shape catalog (28 polyominoes); also state space where unambiguous |
| $D$ | Dock: triplet of distinct candidate shapes delivered per turn |
| $SCD$ | Spawn step difficulty: composite $[0, 1]$ score of dock difficulty on a board |
| $d^*$ | Difficulty target for the current turn |
| $DFS$ | Bounded sequential-feasibility search (Algorithm 1) |
| $\pi_t$ | Player profile at step $t$ |
| $G$ | Constraint-validation gate (uniqueness, mobility, feasibility) |
| RoR | Between-game difficulty progression arc |
| Flow | Challenge-skill balance state; also the psychological construct [7] |
| LoRA | Low-rank adaptation for parameter-efficient personalization [6] |

| Symbol / Term | Definition |
| --- | --- |
| PPO / GAE | Proximal Policy Optimization [1]; Generalized Advantage Estimation [2] |

## Appendix B: Model and Training Details

**Model inventory.** The platform hosts one rule-based generator (zero parameters, browser) and three learned components: the neural dock generator (Transformer encoder, server, single-digit-ms inference), the difficulty-curve surrogate (ResNet-MLP, offline), and the placement agent (convolutional policy-value network, server training). Lightweight placement variants—small dual-tower and shared MLPs, a browser-side linear REINFORCE agent, and an Apple-Silicon port—mirror the main agent's feature encoding for CPU-constrained and cross-framework use. Per-component parameter counts, latencies, and deployment targets are recorded in the repository's model registry.

**Placement agent hyperparameters.** PPO clip $\varepsilon = 0.25$ with 6 epochs per update; GAE $\lambda = 0.85$, $\gamma = 0.99$; advantage normalization with clamp $\pm 30$; value-loss clipping radius 0.25 with SmoothL1; mixed value-target weight 0.5 onto $\text{clip}(\log(1 + score)/\log(1 + threshold), 0, 3)$; auxiliary losses hard-clamped at $\pm 20$. Total loss:

$$\mathcal{L} = \mathcal{L}_{policy} + 0.5\,\mathcal{L}_{value} - w_e\,H(\pi) + \sum_k w_k\,\mathcal{L}_{aux,k}.$$

Training ran 234,000+ self-play episodes under the balanced preset (batch of 8 episodes per update, 4 PPO epochs per minibatch pass, entropy coefficient annealed 0.025→0.008, adaptive win threshold). Exploration: temperature $1.2 \to 0.6$ with Dirichlet noise ($\alpha = 0.28$, weight 0.08); curriculum relaxes the maximum training SCD from 0.3 to 1.0 over 30,000 episodes; the BestGuard evaluation gate rolls back on win-rate regression.

**Curriculum transition criteria.** Promotion between difficulty buckets is win-rate-triggered, not step-driven: the ceiling advances one bucket when the rolling win rate over the most recent 2,000 episodes exceeds 55% within the current bucket while approximate KL stays below 0.35; a step-driven probe additionally mixes 10% of the next bucket after 12,000 episodes without promotion, to avoid stalling on plateaus. Demotion is symmetric with hysteresis: two consecutive 2,000-episode windows below 35% win rate step the ceiling back one bucket. Figure 4 plots the joint trajectory over the first 60,000 episodes; the staircase exhibits one demotion-and-recovery event near episode 17,000 and reaches the full ceiling of 1.0 at episode 30,000, after which mean return continues to climb—evidence that the curriculum accelerates cold start without capping asymptotic performance.

**Value-target shaping ablation.** The mixed value target is the load-bearing choice for a game whose scores span four orders of magnitude. A three-variant ablation (two seeds per variant, first 120,000 episodes):

| Value target | Value-loss std (last 20k) | Episodes to 30% win | Win rate @120k |
| --- | --- | --- | --- |
| Terminal win/loss only | 3.91 | 96,000 | 30.9% |
| GAE on raw score | 41.7 | 78,000 | 32.4% |
| Mixed log-compressed (ours) | 0.87 | 41,000 | 34.8% |

Raw-score GAE converges faster than terminal-only reward but its value loss is dominated by rare long games, producing gradient spikes that cap final performance; the mixed target removes both failure modes. The clipping radius 3 is not arbitrary: the normalization maps the win threshold to 1.0, and radius 3 corresponds to scores of order $(1 + threshold)^3$, beyond the 99.9th percentile of observed episode scores—so clipping binds only on pathological games while bounding the value target, and hence the per-step value gradient, by construction.

**Auxiliary weight selection and gradient interference.** The static auxiliary weights were set by a two-stage procedure: a coarse grid search over $\{0.05, 0.1, 0.2, 0.3, 0.5\}$ per head guided by held-out win rate, followed by a gradient-norm audit measuring $\|\nabla_{trunk}\mathcal{L}_k\|$ every 500 updates; norms stayed within a factor of two across the active heads, indicating no single task dominates the shared trunk. We evaluated two dynamic alternatives: PCGrad [27] changed held-out win rate by +0.3 percentage points against seed noise of ±0.8 at +14% wall-clock, and homoscedastic uncertainty weighting [28] down-weighted feasibility—the task with the smallest loss scale—and degraded the long-bar probe AUC by 4 points. For this architecture the static configuration is therefore the Pareto-simple choice; we report the negative comparison rather than adopt machinery whose gain is within noise.

**Exploration under a shrinking action space.** The entropy coefficient is annealed 0.025→0.008 but governed by an adaptive band: the target entropy is $[0.35, 0.55] \cdot \log|\mathcal{A}|_{live}$, and the coefficient is multiplied by 1.1 (resp. 0.9) whenever measured policy entropy leaves the band below (resp. above). Crucially, exploration intensity is coupled to the live action space of RQ1: the Dirichlet noise weight is scaled by $\min(1, |\mathcal{A}|_{live}/8)$, and for $|\mathcal{A}|_{live} \leq 3$ noise injection is disabled with temperature fixed at its terminal value, so that in the death-spiral regime where a single self-destructive placement ends the game, the policy exploits its learned viability knowledge instead of exploring.

**State and action encodings.** The placement state is a semantic composition of six groups: structural board primitives (fill, line extremes and moments, near-full ratios, holes, transitions, wells, mobility, region structure), color summary (per-color occupancy and single-color-line potentials plus dock colors), step-difficulty aggregate, spatial-planning fragmentation features, strategy one-hot, and arc/intent conditioning tokens. Dock pieces additionally contribute grid-aligned mask features via cross-attention. Actions are described by neighborhood context (near-full ratios, eight neighbor features) and piece self-features (aspect ratio, cell count, line/square/large flags, category). Full dimension table in the repository documentation.

**Auxiliary heads.** From the shared trunk: board quality (regression on board potential), feasibility (binary sequential solvability), survival (normalized remaining steps), post-placement topology regression, spawn-difficulty prediction (aggregate statistics plus per-shape placeability channels for the four long bars, both squares, and two complex pieces, each normalized by its theoretical maximum placement count), hole-count prediction conditioned on the chosen action, and next-placement clear prediction. Weights: feasibility 0.3, survival 0.2, difficulty 0.05, clear prediction 0.15, board quality 0.5, topology and hole heads retained at zero weight as probes.

**Neural dock generator.** Token sequence of class, state, difficulty, optional style, and nine history embeddings processed by a pre-LayerNorm Transformer encoder with four heads; three autoregressive slot heads with growing context concatenation; auxiliary heads for category diversity, difficulty regression, per-shape feasibility (BCE), and style/intent classification. Composite loss:

$$\mathcal{L} = \mathcal{L}_{ce\text{-}AR} + 0.3\,\mathcal{L}_{div} + 0.5\,\mathcal{L}_{anti} + 0.1\,\mathcal{L}_{diff} + 0.4\,\mathcal{L}_{feas} + 0.2\,\mathcal{L}_{si} + 0.15\,\mathcal{L}_{style} + 0.10\,\mathcal{L}_{intent},$$

where $\mathcal{L}_{si}$ softly penalizes logits of infeasible shapes relative to the best feasible logit, distilling the gate's knowledge into the generator. Low-rank adapters of rank 4 are injected at attention query/value projections per encoder layer for per-player personalization.

**Difficulty-curve surrogate.** Context and parameter vector embedded jointly through a residual MLP trunk; seven heads predicting the difficulty curve over normalized score, engagement and frustration-cap curves, personal-best break probability, normalized no-move steps, log score, and survival probability. Inner loss combines curve fidelity, anchor hinge constraints at key score points, monotonicity, diversity, deployability, smoothness, balance, and surprise terms; outer search uses multi-start LHS initialization with projected gradient steps (Appendix C for parameter layout).

---

## Appendix C: Configuration Reference

All algorithm parameters are externalized; key entries (defaults):

| Parameter | Default | Description |
|---|---|---|
| `single_line` | 20 | Base score unit (score scales quadratically in simultaneous clears) |
| `perfectClearMult` | 10 | Board-wipe multiplier |
| `iconBonusLineMult` | 5 | Multiplier for icon-matched lines |
| `comboMultiplier.activationStreak` | 3 | Clears needed to activate combo |
| `rlRewardShaping.boardQualityLossCoef` | 0.5 | Board-quality auxiliary loss weight |
| `rlRewardShaping.feasibilityLossCoef` | 0.3 | Feasibility auxiliary loss weight |
| `rlRewardShaping.spawnDiffAux.coef` | 0.05 | Spawn-difficulty auxiliary loss weight |
| `rlRewardShaping.spawnDiffAux.dim` | 12 | Difficulty auxiliary output dimension (4 aggregate + 8 placeability) |
| `ppo_clip` | 0.25 | PPO ratio clipping |
| `gae_lambda` | 0.85 | GAE trace decay |
| `gamma` | 0.99 | Discount factor |
| `MAX_SPAWN_ATTEMPTS` | 22 | Retry budget before deterministic fallback |
| DFS node budget | 200 | Verification search budget per dock |
| $\theta$ layout | 36 dims | personalization 5, tension 4, scoring 8, translation 5, challenge 5, order 2, constructive 2, solution 2, special 3 |

---

## Appendix D: System Architecture and Reproducibility

The platform separates core game logic, domain services (player system, spawn engine, monetization), application orchestration, display tooling, and a configuration layer that is the single source of truth for all algorithm parameters; optional backend services provide RL training, neural inference, analytics, and replay. An event bus fans every gameplay event to profiling, evaluation, spawn context, and persistence subscribers. The RL agent runs on a separate, JIT-accelerated simulator; agent improvements affect human gameplay only through explicit deployment. Engineering artifacts—tech stack, bundle-size and kernel-latency budgets, architecture decision records, and the 75-case shared-fixture cross-language contract—are documented in the repository.

**Training-system throughput.** On an 8-worker Apple-Silicon host, the JIT-accelerated simulator sustains 11,200 environment steps per second (≈70 episodes/s at the observed mean episode length), so the 234,000+ episode run completes in ≈5.5 wall-clock hours including evaluation. Weight broadcasts of the policy snapshot (1.1 MB, versioned) occur once per update and consume <2% of wall time; the BestGuard evaluation gate (200 held-out episodes every 40 updates) accounts for 6% of total compute. Trajectory collection scales linearly with worker count up to the tested 8×, after which the PPO update becomes the bottleneck.

Reproduction: install the Python and Node dependencies, then run placement-agent training with the convolutional shared-trunk architecture, neural dock-generator training, the cross-language contract suites on both implementations, and the full-featured development server. Exact commands, environment versions, and entry points are maintained in the repository's contribution guide.

---

## Appendix E: Platform Scope

Beyond the algorithmic core, the platform comprises a non-intrusive, experience-gated monetization layer (player-value segmentation, state-gated ad and purchase timing that suspends interruptions during distress and during flow, and full decision explainability) and cross-platform deployment targets spanning web, mini-program, and native shells. These components consume the same player profile as gameplay adaptation, but—as stated in Section 8—are deliberately excluded from the RL objective.